\ifdefined\pdfoutput
  \pdfoutput=1
\fi
\documentclass[10pt,twocolumn]{article}

\usepackage[a4paper,top=1.55cm,bottom=1.7cm,left=1.65cm,right=1.65cm]{geometry}
\usepackage[T1]{fontenc}
\usepackage[utf8]{inputenc}
\usepackage{newtxtext}
\usepackage{newtxmath}
\usepackage{microtype}
\usepackage{inconsolata}
\usepackage{graphicx}
\usepackage{booktabs}
\usepackage{amsmath}
\usepackage{xcolor}
\usepackage{float}
\usepackage{array}
\usepackage[round,authoryear]{natbib}
\usepackage{titlesec}
\usepackage{fancyhdr}
\usepackage{caption}
\usepackage{tcolorbox}
\usepackage{fontawesome5}
\usepackage{enumitem}
\usepackage{hyperref}

\newcommand{\sysname}{FRED}

\providecommand{\Description}[1]{}

\definecolor{accent}{HTML}{075E5A}
\definecolor{accentlink}{HTML}{0B9FAD}
\definecolor{abstractbg}{HTML}{EEF7F6}
\definecolor{rulegray}{HTML}{333333}
\definecolor{muted}{HTML}{555555}

\hypersetup{
  colorlinks=true,
  linkcolor=accent,
  citecolor=accentlink,
  urlcolor=accentlink,
  pdfauthor={Xiao Fan, Hongbin Guo, Yubo Han, Yi Zhang},
  pdftitle={FRED: Frequency-Decorrelated Temporal Ensembles for EEG-fNIRS Imagined-Handwriting Decoding}
}
\setlist[itemize]{leftmargin=1.3em,itemsep=2pt,topsep=3pt}

\titleformat{\section}
  {\large\bfseries\color{accent}}
  {\thesection.}{0.55em}{}
\titleformat{name=\section,numberless}
  {\large\bfseries\color{accent}}
  {}{0pt}{}
\titleformat{\subsection}
  {\normalsize\bfseries\color{accent}}
  {\thesubsection}{0.55em}{}
\titlespacing*{\section}{0pt}{11pt}{5pt}
\titlespacing*{\subsection}{0pt}{8pt}{3pt}

\fancypagestyle{firstpage}{%
  \fancyhf{}
  \fancyfoot[C]{\thepage}
  
}

\newcommand{\preprinttopmatter}{%
\begin{minipage}{\textwidth}
  \thispagestyle{firstpage}
  \noindent
  {\small\bfseries Xidian University}\hfill{\small\itshape ACM MM 2026 BCI Grand Challenge \;\textperiodcentered\; August 2026}\\[-0.35em]
  \color{rulegray}\rule{\textwidth}{0.7pt}\color{black}

  \vspace{0.75em}
  {\fontsize{21.0}{23.5}\selectfont\bfseries\color{accent}
  FRED: Frequency-Decorrelated Temporal Ensembles for\\[-0.05em]
  EEG--fNIRS Imagined-Handwriting Decoding\par}

  \vspace{0.7em}
  {\large\bfseries Xiao Fan,\quad Hongbin Guo,\quad Yubo Han,\quad Yi Zhang\par}
  \vspace{0.22em}
  {\normalsize Xidian University\par}
  {\small\texttt{xdfanxiao@163.com}\par}

  \vspace{0.48em}
  {\small
  \href{https://github.com/XiuFan719/EEG-fNIRS-fuse-method-for-MM-challenge}{\faGithub\ \textbf{Code}}\par}
  \vspace{0.25em}
  \color{rulegray}\rule{\textwidth}{0.7pt}\color{black}

  \vspace{0.72em}
  \begin{tcolorbox}[
    colback=abstractbg,
    colframe=abstractbg,
    boxrule=0pt,
    arc=2.2mm,
    left=3.5mm,right=3.5mm,top=2.8mm,bottom=2.8mm
  ]
  \textbf{Abstract}\quad Imagined handwriting offers a temporally rich paradigm for non-invasive neural decoding, yet reliable recognition across unseen participants remains difficult because scalp EEG is noisy and internally generated stroke sequences vary across individuals. The Multimodal Brain-Computer Interface Grand Challenge provides synchronized EEG and fNIRS for four-class subject-independent handwriting-trajectory classification. We propose FRED, a task-adapted system that models imagined handwriting as a multi-second motor sequence and trains a compact multi-scale temporal network on three complementary EEG frequency views. With three seeds per view, cross-band members produce substantially less-correlated errors than same-band replicas, yielding a clean nine-member ensemble accuracy of 0.8076/0.7242/0.7492 on the public/private/overall test partitions without test-set adaptation or output constraints. The submitted pipeline further incorporates transductive pseudo-label training, three EEG-Conformer members, posterior aggregation, and a paradigm-aware decoder. Because every 12-trial randomization block contains three instances of each class, the final predictions are obtained by Hungarian assignment under the known block quota. On one fixed posterior pool, independent, session-constrained, and block-constrained decoding achieve 0.7600, 0.7758, and 0.7952 overall accuracy, respectively. The complete system reaches 0.8498/0.7718/0.7952, ranking fourth on the private split. A modality audit finds fNIRS-only decoding at chance (0.2511 overall), while adding fNIRS to EEG changes accuracy by only +0.0025. These results identify frequency-diverse temporal EEG modeling and protocol-matched structured inference as the principal sources of performance in this sparse-montage EEG--fNIRS setting. The source code is available at \url{https://github.com/XiuFan719/EEG-fNIRS-fuse-method-for-MM-challenge}.

  \vspace{0.35em}
  {\small\textbf{Keywords:} brain--computer interface; EEG--fNIRS; imagined handwriting; ensemble learning; structured decoding.}
  \end{tcolorbox}
  \vspace{0.55em}
\end{minipage}%
}

\begin{document}
\twocolumn[\preprinttopmatter]

\section{Introduction}
Brain--computer interfaces translate neural activity into communication or control and can restore a degree of interaction for people with severe motor impairment \citep{wolpaw2002bci,chaudhary2016bci}. Recent invasive systems have achieved high-rate decoding of attempted handwriting and speech \citep{handwriting-bci,willett2023high}, but broad deployment depends on non-invasive recordings. Scalp EEG is attractive because it is inexpensive, portable, and has millisecond temporal resolution; however, its low signal-to-noise ratio, volume conduction, and sensitivity to ocular and muscular artifacts make fine-grained decoding considerably more difficult \citep{makeig2012evolving}.

Imagined handwriting is a particularly demanding non-invasive target. Unlike conventional left-versus-right motor imagery, all classes are generated with the same effector and are therefore expected to share much of their gross sensorimotor topography. The discriminative content instead lies in an internally generated sequence of strokes, directions, and transitions that unfolds over several seconds. Neural studies of handwriting implicate both motor-sequence and orthographic processes \citep{handwriting-brain}, while Chinese-character imagery and recent EEG work suggest that the temporal organization of imagined writing remains partially decodable at the scalp \citep{qiu-chinese-writing,ravishankar-imagined-eeg}. These properties motivate two complementary requirements: preserving the within-trial temporal trajectory and accounting for the fact that movement preparation and sensorimotor imagery are expressed over multiple frequency ranges \citep{erd-ers,pfurtscheller2001motor}. The problem becomes harder under subject-independent evaluation, because anatomy, electrode placement, rhythm amplitude, and imagery strategy vary across participants \citep{subject-independent,jayaram2016transfer,saha2020intra}.

The ACM MM 2026 Multimodal Brain-Computer Interface Grand Challenge instantiates this setting with synchronized 27-channel EEG and four-channel fNIRS, four imagined Chinese-character classes, twenty labelled training participants, and ten unseen test participants \citep{challenge}. The multimodal design is motivated by the complementary temporal and spatial properties of EEG and fNIRS \citep{eeg-fnirs-hybrid,hybrid-decision,liu2021review}. Nevertheless, the optical stream in this benchmark contains only HbO/HbR measurements from two locations, and the hemodynamic response evolves much more slowly than the imagined stroke sequence \citep{pinti2020present,scholkmann2014review,yucel2021best}. A suitable solution should therefore exploit the rapid temporal structure of EEG, retain fNIRS without allowing a weak modality to dominate, and remain robust to cross-participant variability.

We propose \sysname{} (Frequency-Decorrelated Temporal Ensembles) to meet these requirements. A multi-scale temporal backbone preserves the evolution of each imagined-writing trial, while three physiologically motivated EEG views (broadband, 4--38\,Hz, and the 8--30\,Hz mu--beta range) provide structured model diversity without assuming a one-character--one-band correspondence. A small fNIRS branch supplies the mandated hemodynamic input. To further adapt the submission to the released test cohort, the system uses high-confidence pseudo-label training and a small cross-architecture Conformer ensemble. Finally, the acquisition protocol is incorporated at inference: each 12-trial block is known to contain three examples of every class, so the aggregated posterior is decoded through an exact quota-constrained assignment rather than independent argmax decisions.

Our contributions are summarized as follows:
\begin{itemize}
  \item We propose \sysname, a subject-independent EEG--fNIRS imagined-handwriting decoder that combines multi-scale temporal modeling, frequency-diverse EEG views, a low-capacity fNIRS branch, and a competition-oriented transductive extension.
  \item We design a frequency-decorrelated ensemble in which cross-band members exhibit markedly lower error correlation than same-band seed replicas; at equal ensemble size, the multi-band pool improves accuracy by 3.5--4.3 percentage points.
  \item We formulate the balanced acquisition protocol as an exact block-level assignment problem and quantify its effect on a fixed posterior pool. The protocol-matched decoder improves overall accuracy from 0.7600 to 0.7952 and contributes to a fourth-place private-set result. A modality audit further quantifies the limited incremental value of the sparse fNIRS stream.
\end{itemize}

\section{Related Work}
\subsection{Imagined Handwriting and Temporal EEG Decoding}
Handwriting combines the planning of sequential hand movements with the retrieval and organization of orthographic forms \citep{handwriting-brain}. Attempted handwriting has enabled high-rate invasive character decoding \citep{handwriting-bci}, motivating efforts to determine how much of this structure remains observable non-invasively. Chinese-character movement imagery has been proposed as a richer motor-imagery paradigm \citep{qiu-chinese-writing}, and recent EEG results indicate that fully imagined handwriting can be decoded above chance despite uncertain internal onset and weak single-trial responses \citep{ravishankar-imagined-eeg}. These findings favor architectures that preserve within-trial dynamics. Compact convolutional models such as EEGNet and ShallowConvNet remain strong EEG baselines \citep{eegnet,shallowconvnet}, while EEG-Conformer combines convolutional tokenization with self-attention for longer context modeling \citep{eeg-conformer}. Our system follows this temporal-decoding line but organizes model diversity around frequency views of the same trial.

\subsection{Subject-Independent EEG Decoding}
Cross-participant decoding is affected by differences in anatomy, electrode placement, signal scale, covariance structure, and task strategy \citep{subject-independent,roy2019deep}. Common remedies include normalization, covariance alignment, and invariant representation learning. Euclidean alignment reduces covariance shift without target labels \citep{alignment}, whereas Riemannian methods classify covariance matrices on their intrinsic geometry \citep{barachant2012multiclass}. Learned objectives pursue a related goal in feature space; supervised contrastive learning, for example, encourages same-class examples from different participants to form compact clusters \citep{supcon}. We retain per-trial normalization and a contrastive term, but use ensemble diversity to address the residual variance that remains among models trained on a small subject cohort.

\subsection{Frequency-Diverse Ensembles}
Motor imagery is expressed over multiple frequency ranges, with mu and beta rhythms closely associated with sensorimotor event-related desynchronization and synchronization \citep{erd-ers,pfurtscheller2001motor}. Filter-bank methods exploit this structure by learning band-specific features and combining them within a classifier \citep{fbcsp}. Deep ensembles reduce predictive variance when their members make complementary errors \citep{deep-ensembles}, yet random-seed replicas of one architecture can remain highly correlated. Our design uses frequency selection as a task-grounded source of diversity: identical temporal architectures are trained independently on complementary EEG bands and combined only in posterior space. This preserves specialization between views and allows their error complementarity to be measured directly.

\subsection{Hybrid EEG--fNIRS and Structured Inference}
Hybrid EEG--fNIRS systems combine fast electrophysiology with slower hemodynamic measurements and have improved motor-imagery decoding under sufficiently rich optical coverage \citep{eeg-fnirs-hybrid,hybrid-attention,buccino2016hybrid,shin2018open}. The gain is protocol dependent because fNIRS must resolve class-relevant spatial or amplitude differences at the timescale supported by the acquisition. The present montage is substantially sparser, motivating an empirical modality audit rather than an assumption of automatic complementarity.

The submitted pipeline also uses two forms of test-time information. High-confidence pseudo-labeling adapts the representation to released unlabelled inputs \citep{pseudolabel,sohn2020fixmatch}. Independently, the known class counts within each randomization block define a structured prediction problem. When class quotas are fixed, the joint maximum-posterior labeling can be recovered by an assignment algorithm rather than trial-wise argmax; we solve this problem with the Hungarian method \citep{hungarian}. The two operations are kept separate so that representation adaptation and protocol-aware inference can be evaluated independently.

\section{Method}
\subsection{Task and notation}
Let the labelled participant set be $\mathcal{S}_{\mathrm{tr}}=\{1,\ldots,20\}$ and the hidden test set be $\mathcal{S}_{\mathrm{te}}=\{21,\ldots,30\}$. The corresponding datasets contain $N_{\mathrm{tr}}=6{,}443$ labelled trials and $N_{\mathrm{te}}=3{,}238$ unlabelled trials during the competition. For trial $i$, the input is $X_i=(X_i^{e},X_i^{n})$, where $X_i^{e}\in\mathbb{R}^{27\times1000}$ is an 8-s EEG segment sampled at 125\,Hz and $X_i^{n}\in\mathbb{R}^{4\times960}$ is a 15-s fNIRS segment sampled at 64\,Hz. The four optical channels are HbO/HbR measurements from two locations, and the target $y_i\in\{1,2,3,4\}$ denotes the imagined character. Accuracy is the official metric.

Each participant contributes repeated 36-trial sessions. Two long rest intervals partition a complete session into three 12-trial randomization blocks. In the labelled cohort, every complete block contains each class exactly three times. The same block boundaries can be reconstructed for the test cohort from trial timestamps alone. The public partition comprises participants 21--23 (972 trials), and the private ranking partition comprises participants 24--30 (2,266 trials). Test labels are used only in the post-competition analyses reported in Section~4.

\begin{figure*}[t]
	\centering
	\IfFileExists{model3.pdf}{%
		\includegraphics[width=0.65\textwidth]{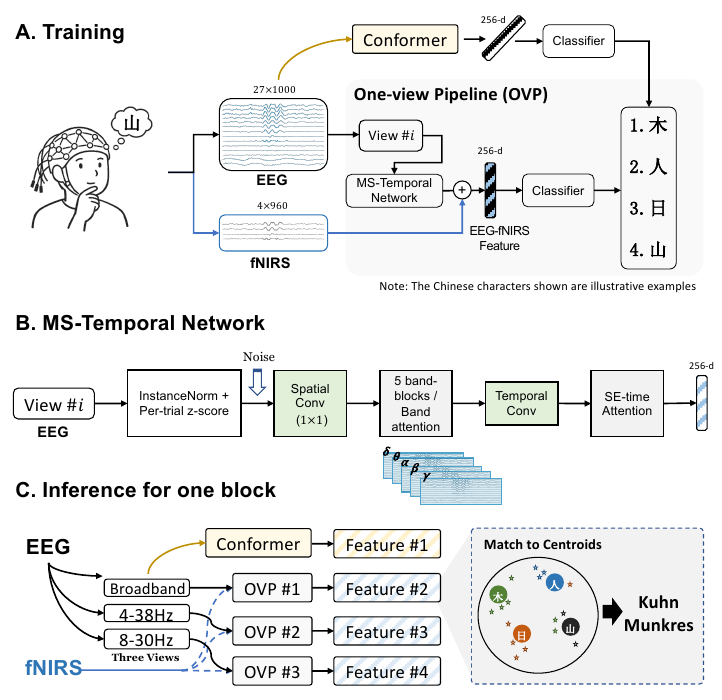}%
	}{%
		\fbox{\parbox[c][5.0cm][c]{0.80\textwidth}{\centering
		\textbf{FRED system overview placeholder}\\[3pt]
		Place \texttt{model3.pdf} in the project directory.}}%
	}
	\Description{Overview of the FRED pipeline: three EEG frequency views and a small fNIRS branch feed temporal models, posterior aggregation, and block-level constrained decoding.}
	\caption{Overview of \sysname. Three EEG frequency views share the same temporal architecture design and are trained with independent parameters. The full system adds pseudo-label training and three EEG-Conformer members. Member posteriors are aggregated before paradigm-aware block decoding.}
	\label{fig:overview}
\end{figure*}

Figure~\ref{fig:overview} summarizes the system. The \emph{clean} configuration is trained only on $\mathcal{S}_{\mathrm{tr}}$, whereas the complete submission additionally uses released unlabelled test inputs and the known block composition.

\subsection{Multi-scale temporal EEG--fNIRS member}
Let $X^{e}\in\mathbb{R}^{27\times T}$ denote the EEG of one trial. To remove participant-specific amplitude scale, each trial is normalized independently,
\begin{equation}
  \tilde{X}^{e} = \mathrm{IN}\!\left(\frac{X^{e}-\mu(X^{e})}{\sigma(X^{e})}\right),
  \label{eq:norm}
\end{equation}
where $\mu,\sigma$ are computed per trial and $\mathrm{IN}$ is instance normalization with learnable affine parameters. Because the model has $\approx\!4.2$M parameters but only 6{,}443 labelled trials, Gaussian noise ($\sigma=0.1$) is added to $\tilde{X}^{e}$ during training only.

A $1\times1$ convolution then mixes electrodes into $C_0=32$ spatial maps, $H_0=W_{s}\ast\tilde{X}^{e}$, after which five parallel temporal blocks with different kernel lengths $k_b$ extract patterns at different temporal scales,
\begin{equation}
  H_b = \phi\!\left(W_b \ast H_0\right),\qquad
  k_b \in \{31,21,15,11,7\},
  \label{eq:bands}
\end{equation}
with $\phi$ a nonlinearity. The kernel lengths are ordered from long to short and are labelled by the canonical rhythm names ($\delta,\theta,\alpha,\beta,\gamma$) for readability; they are multi-scale temporal filters, not exact band-pass decompositions, and the explicit frequency selection is performed at the input (Section~\ref{sec:ensemble}). Their outputs are combined by a learned attention over scales,
\begin{equation}
  \alpha_b=\frac{\exp(w_b^{\top}\bar{H}_b)}{\sum_{b'}\exp(w_{b'}^{\top}\bar{H}_{b'})},
  \qquad
  H=\sum_{b=1}^{5}\alpha_b H_b,
  \label{eq:bandattn}
\end{equation}
where $\bar{H}_b$ is the time-averaged descriptor of $H_b$.

Two residual temporal blocks widen the representation ($32\!\to\!64$), followed by squeeze-and-excitation over the \emph{time} axis, which reweights informative time steps,
\begin{equation}
  \hat{H} = H \odot \sigma\!\left(W_2\,\phi(W_1 \, \mathrm{pool}_c(H))\right),
  \label{eq:se}
\end{equation}
with $\mathrm{pool}_c$ pooling over channels and $\sigma$ the logistic function. The tensor $\hat{H}$ is vectorized without global temporal pooling, thereby retaining the within-trial ordering represented along the temporal axis.

The fNIRS branch is intentionally small. Dilated one-dimensional convolutions encode the four slow hemodynamic channels into $g^{n}$, which is concatenated with the flattened EEG feature before the head,
\begin{equation}
  g = \left[\,\mathrm{vec}(\hat{H});\; g^{n}\,\right] \in \mathbb{R}^{256}.
  \label{eq:fuse}
\end{equation}
The branch is kept low-capacity to limit the influence of the sparse hemodynamic input; its contribution is quantified in Section~\ref{sec:fnirs}.

The submitted temporal members discard the linear head at inference and read out $g$ by cosine nearest-class-centroid matching, following the prototype-based view of class representations \citep{prototypical}. Prototypes are the class means over all labelled participants, $m_c=\frac{1}{|\mathcal{I}_c|}\sum_{i\in\mathcal{I}_c} g_i$, and the posterior is
\begin{equation}
  p(c\mid g)=\frac{\exp\!\big(\tau\,\cos(g,m_c)\big)}
                    {\sum_{c'}\exp\!\big(\tau\,\cos(g,m_{c'})\big)},
  \qquad \tau=10 .
  \label{eq:ncc}
\end{equation}
The Conformer members instead use their learned classification heads. A post-hoc comparison found no stable independent advantage for centroid readout, so this choice is reported as an implementation detail rather than a contribution.

Members are trained with cross-entropy on the linear head together with a supervised contrastive term on $g$ \citep{supcon},
\begin{equation}
  \mathcal{L}=\mathcal{L}_{\mathrm{CE}}
  + \lambda\,\mathcal{L}_{\mathrm{SupCon}}(g;\,\text{temp.}=0.1),
  \qquad \lambda=1.0 .
  \label{eq:loss}
\end{equation}
The contrastive term encourages same-class examples from different participants to form a more compact representation. Models are optimized for 26 epochs with Adam and weight decay $2\times10^{-3}$; each configuration uses three random seeds, 42, 7, and 99. Each frequency view is trained \emph{independently}: the members share an architecture, not weights, and no loss couples them.

\subsection{Frequency-decorrelated temporal ensemble}
\label{sec:ensemble}
We generate three in-model FIR-filtered EEG views:
\begin{equation}
  \mathcal{B}=\{0.5\text{--}40,\;4\text{--}38,\;8\text{--}30\}\ \mathrm{Hz}.
\end{equation}
The broadband view preserves the full task-related spectrum, the 4--38\,Hz view suppresses slow drift and high-frequency contamination, and the 8--30\,Hz view focuses on the canonical mu--beta sensorimotor range. We train the same temporal architecture separately for every view and seed, yielding nine members. Their posteriors are aggregated geometrically. For $M$ members, the ensemble posterior is
\begin{equation}
  \bar p(c\mid x)=
  \frac{\exp\left(\frac{1}{M}\sum_{m=1}^{M}\log(p_m(c\mid x)+\epsilon)\right)}
       {\sum_{c'}\exp\left(\frac{1}{M}\sum_{m=1}^{M}\log(p_m(c'\mid x)+\epsilon)\right)},
  \label{eq:geom}
\end{equation}
with a small $\epsilon$ for numerical stability. Averaging in log-probability space reduces the influence of an isolated over-confident member and provides the score matrix used by the assignment decoder. The method does not require a fixed mapping from characters to bands; its purpose is to create complementary decision errors under cross-participant variability.

\subsection{Transductive extension}
The test participants and their unlabelled inputs were released for prediction, so we also used a transductive extension. The current ensemble assigns pseudo-labels to test trials, retains predictions with confidence above 0.90, and retrains the three frequency-view models with these additional examples. The pseudo-label set is refreshed after each round; low-confidence trials remain unlabelled. No test ground-truth label enters training. Because no test labels are available to select a round, we monitor two label-free diagnostics: (i) the predicted class histogram, which should remain close to the balanced protocol, and (ii) round-to-round prediction agreement. We retain the third round, after which the histogram begins to drift and agreement no longer tightens. The final pool contains nine pseudo-label-trained temporal members and three EEG-Conformer members \citep{eeg-conformer}, providing architectural diversity in addition to frequency diversity. All outputs are calibrated as four-class posteriors and combined through Eq.~\eqref{eq:geom} before structured decoding.

\subsection{Paradigm-aware block decoding}
Trial timestamps reveal that each 36-trial session is divided by two long rest intervals into three 12-trial blocks. In every complete training block, each of the four characters occurs exactly three times; this holds for all 536 complete blocks. Let $p_{ic}$ be the aggregated posterior for trial $i$ and class $c$ in a block. Instead of independent prediction, we solve
\begin{equation}
  \max_{z_{ic}}\ \sum_{i=1}^{12}\sum_{c=1}^{4} z_{ic}\log p_{ic},
  \quad
  \sum_c z_{ic}=1,\quad \sum_i z_{ic}=3,
  \label{eq:assignment}
\end{equation}
where $z_{ic}\in\{0,1\}$. We expand each class into three assignment slots and solve Eq.~\eqref{eq:assignment} exactly with the Hungarian algorithm. The partition is reconstructed from timestamps alone: the two longest within-session rest gaps define the three blocks, and the same procedure is applied to training and test sessions. For an incomplete block, feasible class-count vectors summing to the observed number of trials are enumerated and the highest-scoring assignment is retained. A coarser session constraint $[9,9,9,9]$ is evaluated as a controlled comparison. Reapplying the reconstructed blocks and assignment code to the cached final posterior reproduces the official 0.84979 submission exactly, providing a direct verification of the post-competition analysis.

\section{Experiments}
\subsection{Evaluation protocol}
During development, we used four subject-disjoint folds: $\{1\!:\!5\}$, $\{6\!:\!10\}$, $\{11\!:\!15\}$, and $\{16\!:\!20\}$. Every labelled participant is therefore predicted only by a model trained on the other fifteen participants. The public test partition contains participants 21--23 (972 trials), and the private ranking partition contains participants 24--30 (2,266 trials). Overall accuracy is computed over all 3,238 test trials. All analyses using the final test labels were conducted after the submitted prediction and ranking were fixed.

\subsection{Reporting levels and post-competition audit}
We report three performance levels. The \emph{clean} ensemble uses only the twenty labelled participants; the \emph{transductive posterior} additionally uses pseudo-label training and Conformer members but predicts each trial independently; the \emph{submitted result} further applies the block quota. This separation prevents the final score from being interpreted as the accuracy of a standalone trial classifier.

The public leaderboard was available during the competition, whereas complete test labels were released only after the ranking was fixed. The frequency, decoder, participant, and modality analyses are therefore retrospective audits that reuse cached predictions and do not retrain or select a new submission. Confidence intervals use participant-level bootstrap resampling, preserving dependence among trials from the same participant.

\subsection{Implementation details}
\label{sec:impl}
Each temporal member has $\approx\!4.2$M parameters and is trained for 26 epochs with Adam on a single GPU; one member takes roughly 20 minutes, so the nine clean members, the nine pseudo-label members, and the three Conformer members amount to about seven GPU-hours in total. Inference is inexpensive: the member posteriors for all 3{,}238 test trials are produced in a few minutes, and the Hungarian assignment is negligible ($270$ blocks, each a $12\times12$ problem solved in $O(n^{3})$). All decoding comparisons in Section~\ref{sec:decode} reuse one cached posterior pool, so they differ only in the decision rule. Code, model configurations, prediction files, and the scoring scripts for the public/private/overall partitions will be released with the paper.

\subsection{System performance and ranking}
Table~\ref{tab:system} reports cumulative configurations of the submitted pipeline. The tuned single-view baseline reaches 0.7153 overall. The clean three-band ensemble raises this to 0.7492 without pseudo-label adaptation or quota decoding, with a larger gain on the seven private participants than on the public participants. Pseudo-label training and the cross-architecture pool provide smaller additional gains, and the full submitted system reaches 0.7952 overall and 0.7718 on the private split. The clean ensemble is the directly transferable subject-independent decoder; the later stages are specific to the released test cohort and the known acquisition protocol. Their results are therefore reported separately rather than attributed to the trial-level model.

\begin{table}[t]
\caption{Cumulative accuracy of the main submitted-system stages. ``Clean'' excludes pseudo-label adaptation and output quotas. Rows are system variants, not single-factor causal ablations.}
\label{tab:system}
\centering
\small
\setlength{\tabcolsep}{3.6pt}
\begin{tabular}{lccc}
\toprule
System stage & Public & Private & Overall \\
\midrule
Single-view temporal model & .7829 & .6862 & .7153 \\
Clean frequency ensemble & .8076 & .7242 & .7492 \\
+ pseudo-label training & .8117 & .7317 & .7557 \\
+ Conformer and aggregation & .8272 & .7471 & .7712 \\
+ block decoding (submitted) & \textbf{.8498} & \textbf{.7718} & \textbf{.7952} \\
\bottomrule
\end{tabular}
\end{table}

The official private leaderboard is shown in Table~\ref{tab:ranking}. \sysname{} was submitted by team CBI and ranked fourth, only 0.0013 behind the third-ranked team and 0.0190 ahead of the fifth-ranked team.

\begin{table}[t]
\caption{Top five teams on the official private ranking partition.}
\label{tab:ranking}
\centering
\small
\begin{tabular}{clc}
\toprule
Rank & Team & Private accuracy \\
\midrule
1 & HappyBCI & .8041 \\
2 & whiteram & .8010 \\
3 & AI4Art & .7732 \\
4 & CBI (\sysname) & .7718 \\
5 & PR-EEG & .7529 \\
\bottomrule
\end{tabular}
\end{table}

\subsection{Why frequency diversity helps}
\label{sec:freq}
The clean multi-band gain could arise either because band-limited members are individually stronger or because they make complementary errors. The controlled analysis supports the second explanation. Across seed replicas of the same band, the mean pairwise error correlation is 0.82; across different frequency views it falls to 0.29. At equal ensemble size, a multi-band pool exceeds a broadband-only pool by 0.035 accuracy for three members (95\% subject-bootstrap CI $[0.013,0.055]$) and by 0.043 for six members ($[0.026,0.059]$). Both intervals exclude zero. Individual band members are not uniformly better than broadband members; the improvement appears after aggregation.

\begin{table}[t]
\caption{Controlled frequency-diversity analysis. Error correlation is computed between member-wise correctness vectors; accuracy differences compare equal-size multi-band and broadband-only pools.}
\label{tab:frequency}
\centering
\small
\begin{tabular}{lc}
\toprule
Measure & Result \\
\midrule
Same-band seed error correlation & .82 \\
Cross-band error correlation & .29 \\
Multi-band $-$ broadband, 3 vs. 3 & +.035 $[.013,.055]$ \\
Multi-band $-$ broadband, 6 vs. 6 & +.043 $[.026,.059]$ \\
\bottomrule
\end{tabular}
\end{table}

These results justify the term \emph{frequency-decorrelated}: the useful diversity is tied to the frequency view, not merely to random initialization. The effect is consistent at two ensemble sizes, which makes it unlikely to be a single-seed artifact. At the same time, the analysis is predictive rather than neurophysiological: it shows that the views make different mistakes, not why a specific trial is resolved by a specific band. In particular, it does not establish that each character has a unique frequency signature, and we make no such claim.

\subsection{Effect of the protocol constraint}
\label{sec:decode}
To isolate decoding from model changes, Table~\ref{tab:decode} applies three decoders to one fixed posterior pool. Both constraints improve independent argmax on every split. Session-level balancing adds 0.0158 overall, while the block-level constraint adds 0.0352 and reaches the submitted 0.7952 score. At the participant level, block minus session is 0.0195 with a bootstrap 95\% CI of $[0.011,0.028]$; block minus argmax is 0.035 $[0.024,0.045]$; and session minus argmax is 0.016 $[0.009,0.024]$.

\begin{table}[t]
\caption{Controlled decoding comparison on one fixed posterior pool.}
\label{tab:decode}
\centering
\small
\begin{tabular}{lccc}
\toprule
Decoder & Public & Private & Overall \\
\midrule
Independent argmax & .8107 & .7383 & .7600 \\
Session quota $[9,9,9,9]$ & .8282 & .7533 & .7758 \\
Block quota $[3,3,3,3]$ & \textbf{.8498} & \textbf{.7718} & \textbf{.7952} \\
\bottomrule
\end{tabular}
\end{table}

Compared with argmax, block decoding corrects 235 predictions and breaks 121, whereas session decoding corrects 159 and breaks 108. The block prior therefore corrects almost twice as many trials as it harms. It is not a replacement for neural evidence: the two hardest participants, 28 and 30, obtain net gains of only one and three trials. The constraint is most useful when the posterior is already informative but locally imbalanced.

\subsection{Participant-level heterogeneity}
\label{sec:persubject}
\begin{figure}[t]
  \centering
  \IfFileExists{persubject.pdf}{%
    \includegraphics[width=\linewidth]{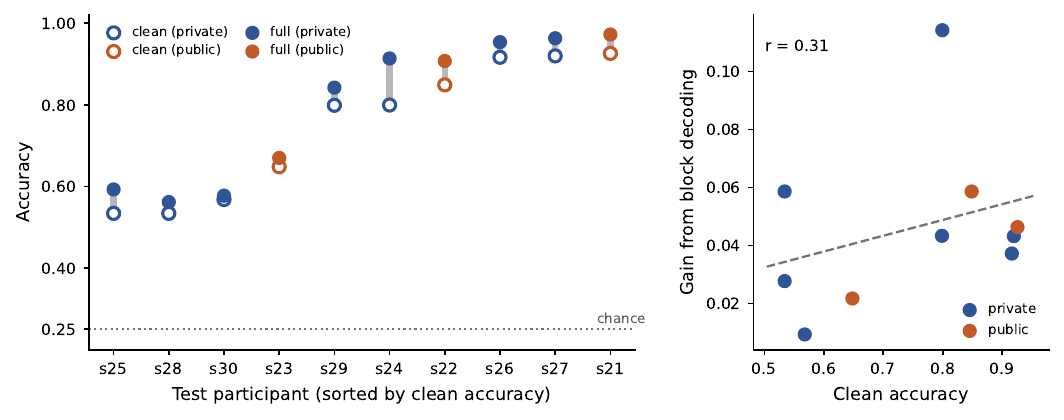}%
  }{%
    \fbox{\parbox[c][3.2cm][c]{0.94\linewidth}{\centering
      \textbf{Per-participant figure placeholder}\\[3pt]
      Replace with \texttt{persubject.png}.}}
  }
  \Description{Two plots showing clean and full-system accuracy for ten test participants and full-minus-clean gain as a function of clean accuracy.}
  \caption{Per-participant results on the true test labels. \emph{Left:} clean (open) and full (filled) accuracy, ordered by clean accuracy. \emph{Right:} full-minus-clean gain against clean accuracy. Accuracy varies markedly across participants: six have clean accuracy .799--.926 (group mean .868; full .925), while four (23, 25, 28, and 30) have clean accuracy .534--.648 (group mean .571; full .600).}
  \label{fig:persubject}
\end{figure}

Figure~\ref{fig:persubject} orders the ten test participants by clean-system accuracy. Six are decoded at .799--.926, whereas four (23, 25, 28, and 30) fall between .534 and .648; three of these four belong to the private partition, explaining much of the split gap. The full pipeline improves the higher-accuracy group by .057 on average and the lower-accuracy group by .029. Together with the fixed-posterior audit, this indicates that protocol constraints refine informative posteriors but cannot compensate for weak trial-level representations.

\subsection{Contribution of fNIRS}
\label{sec:fnirs}
Table~\ref{tab:modality} evaluates fold-averaged softmax outputs on the true test labels. EEG-only decoding reaches 0.7128 overall, whereas fNIRS-only decoding remains at chance (0.2511). Fusion changes overall accuracy by only +0.0025 and has opposite signs on the public and private partitions. On EEG errors, fNIRS assigns 0.2496 probability to the true class, essentially the uniform value; the observed oracle accuracy (0.7863) is also close to the 0.7846 expected when a four-class chance predictor is paired with the EEG model.

\begin{table}[t]
\caption{Test-set modality audit (chance $=.25$). The fourth row reports the mean fNIRS probability assigned to the true class on trials missed by EEG.}
\label{tab:modality}
\centering
\small
\setlength{\tabcolsep}{4pt}
\begin{tabular}{lccc}
\toprule
Decoder / statistic & Public & Private & Overall \\
\midrule
EEG only & .7798 & .6840 & .7128 \\
fNIRS only & .2562 & .2489 & .2511 \\
EEG + fNIRS & .7757 & .6893 & .7153 \\
\midrule
fNIRS $P(y)$ on EEG errors & .2499 & .2495 & .2496 \\
Oracle: EEG or fNIRS correct & .8333 & .7661 & .7863 \\
\bottomrule
\end{tabular}
\end{table}

The limited contribution is plausible because four channels at two locations provide sparse coverage, while the hemodynamic response evolves over seconds and can blur fine within-trial ordering or overlap across nearby trials \citep{eeg-fnirs-hybrid,hybrid-attention,pinti2020present,scholkmann2014review}. These are possible explanations rather than mechanisms established here. The supported conclusion is narrower: under this montage, protocol, and evaluated decoders, fNIRS provides little measurable trial-level information beyond EEG.

\section{Conclusion}
We proposed \sysname, a subject-independent EEG--fNIRS imagined-handwriting system combining multi-scale temporal modeling, frequency-diverse EEG ensembling, transductive adaptation, and protocol-aware inference. The submission achieves 0.7952 overall accuracy and 0.7718 on the private partition, placing fourth. Controlled analyses identify two main sources of performance: cross-band members make substantially less-correlated errors than same-band replicas, and a quota applied at the true 12-trial randomization unit improves a fixed posterior more than independent prediction or session balancing. The clean ensemble is reported separately because pseudo-label training is transductive and block decoding requires a known balanced unit. The fNIRS finding is likewise specific to the four-channel montage and evaluated protocol. Within these limits, the results support preserving the temporal motor sequence, diversifying EEG evidence through complementary frequency views, and applying design priors only at the unit where they are guaranteed.


\bibliography{FRED_refs_fullwidth}

\end{document}